\documentclass[sn-mathphys-num]{sn-jnl}

\usepackage{graphicx}%

\usepackage{amsmath,amssymb,amsfonts}%
\usepackage{amsthm}%
\usepackage{mathrsfs}%
\usepackage[title]{appendix}%
\usepackage{xcolor}%
\usepackage{textcomp}%
\usepackage{manyfoot}%
\usepackage{booktabs}%
\usepackage{algorithm}%
\usepackage{algorithmicx}%
\usepackage{algpseudocode}%
\usepackage{listings}%
\usepackage{hhline}
\usepackage{multirow}%

\theoremstyle{thmstyleone}%
\theoremstyle{thmstyletwo}%

\theoremstyle{thmstylethree}%

\begin{document}

\title[Shared Latent Space by Both Languages for Non-Autoregressive Neural Machine Translation]{Shared Latent Space by Both Languages for Non-Autoregressive Neural Machine Translation}


\author[1]{\fnm{DongNyeong} \sur{Heo}}\email{dnheo@handong.ac.kr}

\author*[1]{\fnm{Heeyoul} \sur{Choi}}\email{hchoi@handong.edu}


\affil[1]{\orgdiv{Computer Science and Electrical Engineering}, \orgname{Handong Global University}, \orgaddress{\street{558 Handong-ro}, \city{Pohang}, \postcode{37554}, \state{Gyeongbuk}, \country{Republic of Korea}}}




\abstract{Non-autoregressive neural machine translation (NAT) offers substantial translation speed up compared to autoregressive neural machine translation (AT) at the cost of translation quality. Latent variable modeling has emerged as a promising approach to bridge this quality gap, particularly for addressing the chronic multimodality problem in NAT. In the previous works that used latent variable modeling, they added an auxiliary model to estimate the posterior distribution of the latent variable conditioned on the source and target sentences. However, it causes several disadvantages, such as redundant information extraction in the latent variable, increasing the number of parameters, and a tendency to ignore some information from the inputs. In this paper, we propose a novel latent variable modeling that integrates a dual reconstruction perspective and an advanced hierarchical latent modeling with a shared intermediate latent space across languages. This latent variable modeling hypothetically alleviates or prevents the above disadvantages. In our experiment results, we present comprehensive demonstrations that our proposed approach infers superior latent variables which lead better translation quality. Finally, in the benchmark translation tasks, such as WMT, we demonstrate that our proposed method significantly improves translation quality compared to previous NAT baselines including the state-of-the-art NAT model.}




\maketitle

\section{Introduction}
\label{sec:introduction}

Neural machine translation (NMT) has gained considerable attention over the past decade \cite{bahdanau2014neural, vaswani2017attention}, 
with notable advancements achieved through autoregressive decoding architectures. However, such architectures are constrained by their sequential nature, as the decoder must iteratively generate each target word based on previously generated context, thereby limiting translation speed. In response to this constraint, the non-autoregressive NMT (NAT) paradigm emerged, aiming to generate entire target words without reliance on prior context. Notably, NAT models have demonstrated significantly enhanced translation speeds compared to their autoregressive counterparts (AT) \citep{gu2017non, lee2018deterministic, libovicky2018end, shu2020latent, gu2020fully, bao2022latent, huang2022directed, ma2023fuzzy, gui2024non}.

However, non-autoregressive translation (NAT) models frequently make a trade-off between translation quality and the acceleration of translation speed. The primary factor contributing to the diminished translation quality is the `multimodality problem', which arises when specifying a mode for the target sentence becomes challenging due to the incorporation of distributions across all constituent words, particularly in cases of multi-modal distribution within the target sentence. This challenge stems from the fact that the NAT model computes the output distribution of a word without conditioning on previous target words. Several methods have been proposed to address this issue, including dataset preprocessing \citep{kim2016sequence, gu2017non}, iterative decoding processes \citep{lee2018deterministic}, and modifications to the objective function \citep{libovicky2018end}. Additionally, the introduction of a latent variable that implicitly represents a mode of the target sentence presents a promising approach to mitigating this problem, often resulting in performance improvements \citep{shu2020latent, bao2022latent, gu2020fully}.

The conventional methodology for latent variable modeling entails incorporating an auxiliary model alongside the encoder-decoder architecture. This auxiliary model is utilized to approximate the posterior distribution, $q(Z|X,Y)$, of the latent variable, $Z$, conditioned on both the source sentence, $X$, and the target sentence, $Y$. The typical encoder and decoder components estimate the prior distribution, $p(Z|X)$, and the likelihood, $p(Y|Z,X)$, respectively. Subsequently, the entire model architecture is trained using the evidence lower-bound objective (ELBO) function, akin to the framework employed in variational auto-encoder (VAE) models \citep{kingma2013auto}.

Nevertheless, this latent variable modeling approach suffers from three primary drawbacks. Firstly, the latent variable is solely utilized to estimate the likelihood distribution of the target sentence during the training phase. Consequently, the latent variable may encode target language-specific information, which is absent during testing and proves ineffective for the translation task. Secondly, the inclusion of the auxiliary posterior model significantly increases the number of parameters, so that it can cause practical issues, such as hardness in implementing as on-device application. Lastly, the auxiliary posterior model usually possesses the ability to manipulate the ratio of extracted information from each input. Consequently, there is the possibility that the posterior model might disregard either the source or target sentence based on the log-likelihood or the Kullback-Leibler (KL) divergence terms within the ELBO function, respectively. We refer to this final drawback as the one-sided posterior collapse (OSPC) problem. With the OSPC problem, the model's performance during testing may suffer, similar to the first drawback, if it has disregarded the source sentence during training. On the other hand, the model's latent variable architecture may lose its meaning if the target sentence has been disregarded during training.

\begin{figure}
    \centering
    \includegraphics[width=0.4\linewidth]{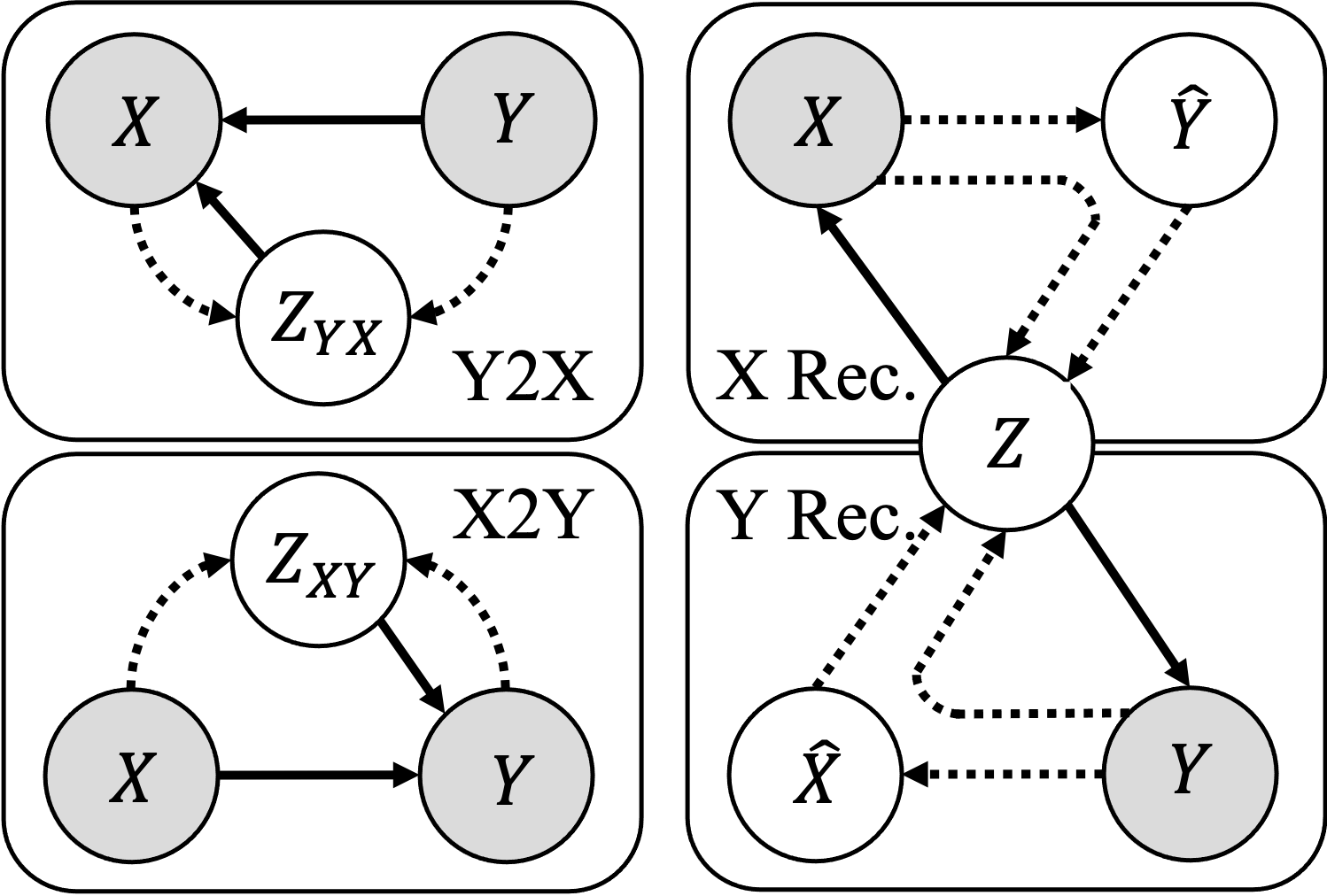}
    \caption{Graphical models of the conventional latent variable model (left)
    white circles represent observation and latent variables, respectively, while solid and dashed lines denote generation and inference processes, respectively. In contrast to prior approaches, we conceptualize the entire translation systems as dual reconstructions, with which we propose dual hierarchical latent variable model, so that the intermediate latent variable, $Z$, is inferred by the observation and the deeper latent variable (translated sentence). Furthermore, we share the intermediate latent variable across the reconstruction processes of both languages.
    }
    \label{fig:graphical_model_comparison}
\end{figure}

This paper introduces a novel latent variable model designed to alleviate or circumvent the aforementioned limitations. Our methodology is rooted in the concept of dual reconstruction perspective, which posits the translated target sentence as an aligned latent variable of the source sentence (observed variable) in the example of source sentence reconstruction \citep{he2016dual, heo2022end}.
It interprets source-to-target and target-to-source translations as inference and generation, respectively, constituting the reconstruction of the source sentence. Analogously, the reconstruction process of the target sentence follows a similar framework but with reversed perspectives. Between the observed and latent variables, we introduce an additional intermediate latent variable, establishing a dual hierarchical latent variable model. A pivotal characteristic of this intermediate latent variable is that it is shared across the reconstruction processes of both languages. Fig. \ref{fig:graphical_model_comparison} illustrates the distinction between our proposed dual hierarchical latent variable model and the conventional approach. While the conventional approach entails separate translation processes with distinct latent variables, our approach shares the intermediate latent variable that is inferred and used to generation by both reconstruction processes.

The shared intermediate latent variable facilitates the simultaneous generation of both source and target sentences, thereby promoting the extraction of language-agnostic information from both source and target sentences. Furthermore, guided by the dual reconstruction perspective, we leverage the two encoders, which serve as an inference and a generation models within a reconstruction process, to jointly infer the posterior distribution of the shared intermediate latent variable. This approach resembles the advanced inference technique (ladder inference) utilized in ladder VAE (LVAE), where both inference and generation models collaborate to infer the posterior distribution \citep{sonderby2016ladder}. Therefore, we call our proposed model as {\em LadderNMT}. Notably, this characteristic eliminates the need for an auxiliary posterior model in our proposed latent variable modeling, resulting in parameter savings. Remarkably, this ladder inference mechanism substantially mitigates the OSPC problem, as it explicitly incorporates both source and target information through a simple function.

In our experiment results, we demonstrate that our LadderNMT surpasses the translation performance of previous methodologies, notably outperforming a state-of-the-art NAT model, FullyNAT \citep{gu2020fully}, while achieving an average reduction of approximately 30\% in parameters. Furthermore, through qualitative analyses conducted on the intermediate latent space, we illustrate the promising regularization effect of our LadderNMT in extracting language-agnostic information. Lastly, experimental investigations reveal the presence of the OSPC problem in conventional approaches, while our LadderNMT effectively circumvents this issue.

\section{Related Works}
\label{sec:related_works}

\subsection{Variational NMTs}
\label{subsec:variationl_nmt}
The task of NMT involves mapping from source sentences to target sentences, which can be understood as a many-to-many mapping. This is exemplified by instances where a single source sentence may correspond to multiple potential target sentences, influenced by factors such as stylistic preferences and intended communicative purposes. However, conventional decoding strategies, reliant solely on deterministically encoded representations of source sentences, may struggle to effectively handle this nuanced mapping. In response to this challenge, variational NMT (VNMT) was introduced \citep{zhang2016variational, su2018variational}. VNMT addresses this by estimating the likelihood of target sentences given source sentences, denoted as $p(Y|X)$, through marginalization with respect to a latent variable $Z$, as depicted by the equation:
\begin{align}
    p(Y|X) = \int_{Z} p(Y,Z|X)dZ = \int_{Z} p(Y|Z,X)p(Z|X)dZ, \nonumber
\end{align}
where $X=\{x_1,x_2,...,x_{T_x}\}$ is a source sentence consisting of $T_x$ source words $x \in V_x$, and $V_x$ is the vocabulary set of the source language. Similarly, $Y=\{y_1,y_2,...,y_{T_y}\}$ is a target sentence with $T_y$ target words $y \in V_y$. $Z=\{z_1,z_2,...,z_{T_z}\}$ is the set of latent variables with $T_z$ vectors $z \in \mathbb{R}^{d_z}$.
Notably, the stochastic property of sampling $Z$ facilitates the representation of multi-modal translations. To circumvent sampling from the computationally expensive true posterior distribution, VNMT employs a variational posterior distribution $q(Z|X,Y)$ to approximate $p(Z|X,Y)$.
Finally, the objective is designed by ELBO function as follows: 
\begin{align}
    \mathcal{J}_{vnmt}(X,Y) &= \mathbb{E}_{q(Z|X,Y)} [\log{p(Y|Z,X)}] - \alpha D_{KL}[q(Z|X,Y) \parallel p(Z|X)], \label{eq:elbo_original}
\end{align}
where $D_{KL}$ means Kullback-Leibler divergence (KL). 
The likelihood and prior distributions, $p(Y|Z,X)$ and $p(Z|X)$, are parameterized by encoder and decoder models of the NMT, respectively. The approximated posterior distribution $q(Z|X,Y)$ is typically parameterized by another encoder-decoder model. During testing, $Z$ is sampled from the prior distribution $p(Z|X)$ since the target sentence is unavailable, necessitating minimizing the KL divergence term to align the prior and posterior distributions. $\alpha$ is the coefficient that manipulates the effect of KL divergence.

This paper addresses several challenges within this framework. Firstly, the latent variable in the ELBO equation is solely utilized for computing the target sentence likelihood, potentially leading the model to capture target language-specific information during training, which is both undesirable and unavailable during testing. Secondly, the auxiliary posterior model introduces a considerable number of parameters, typically employing Transformer architecture \citep{vaswani2017attention}. Lastly, a high-complexity model combining with an imbalance in importance between the two objective terms may lead to one-sided information extraction between source and target sentences, termed as the one-sided posterior collapse (OSPC) problem. For instance, if the importance of the likelihood term is greater, then the latent variable may preserve the information from the target sentence while disregarding the source sentence information. On the other hand, if the importance of the KL divergence term is greater, the posterior distribution may fully depend on the source sentence to achieve the optimal solution that minimizes KL divergence: $q(Z|X,Y)=p(Z|X)$. This paper includes investigating the existence of the OSPC problem in conventional latent variable models.

Since the advent of VNMT, several advancements have emerged \citep{su2018variational, shah2018generative, zheng2019mirror}.
Notably, some prior studies proposed to share the latent space between the source and target sentences, leveraging this shared latent variable to estimate the likelihoods of NMT and language model (LM), denoted as $p(Y|Z,X)$ and $p(X|Z)$, respectively \citep{shah2018generative}. Particularly, MGNMT \citep{zheng2019mirror}, which is an auto-regressive model, proposed a concurrent execution of both NMT and LM tasks on source and target sides. These approaches demonstrated that their usages of latent variable to generate both $X$ and $Y$ can encapsulate language-agnostic information, thereby enhancing translation performance.
However, despite these advancements, the issue of reliance on an auxiliary posterior model still persists. In addition, those previous works consider both source and target sentences are observation variables that are simultaneously aligned with the latent variable, so that they employ the standard Gaussian distribution for prior distribution without conditioning on an input sentence: $p(Z)$ instead of $p(Z|X)$ in Eq. \ref{eq:elbo_original}. It provides uninformative regularization of the KL divergence term, and it may necessitate numerous iterative steps to refine the latent variable from random noise during testing. In contrast, our proposed dual hierarchical latent variable model, built upon LAVE's ladder inference methodology, obviates the need for an auxiliary posterior model and allows the employment of the prior distribution conditioned on an input sentence. Therefore, it reduces parameters and facilitates a superior quality of the initial latent variable sampled from the conditional prior distribution during testing.

\subsection{Non-Autoregressive NMTs (NATs)}
\label{subsec:non-autoregressive_nmt}
Utilizing the latent variable modeling, the autoregressive decoding process of NMT models follows the formulation:
\begin{align}
    p(Y|Z,X) = \prod_{i=1}^{T_y} p(y_i|Z,X,Y_{<i}). \nonumber
\end{align}
The translation speed decreases proportionally with the length of the sentence due to the conditional probability in the right-hand side multiplication conditioned to the previously generated target words.

To enhance the speed, non-autoregressive decoding was introduced \citep{gu2017non}. The conditional probability of each target word is independent of the preceding target words:
\begin{align}
    p(Y|Z,X) = \prod_{i=1}^{T_y} p(y_i|Z,X). \nonumber
\end{align}
Consequently, the decoding process can be executed fully in parallel, leading to significant speed improvements over autoregressive approaches. Since its inception, non-autoregressive translation (NAT) has garnered considerable research attention, resulting in a plethora of methodologies \citep{lee2018deterministic, libovicky2018end, shu2020latent, gu2020fully, huang2022directed, bao2022latent, ma2023fuzzy, gui2024non}.

However, the increased translation speed often comes at the cost of translation quality due to the lack of dependencies between target words, leading to the `multimodality problem' \citep{gu2017non}. Latent variable modeling within the VNMT framework has emerged as a promising solution to this challenge, as it can implicitly represent a unified mode for the entire target sequence decoding. The NAT approaches that use the latent variable modeling of the VNMT framework, such as LaNMT \citep{shu2020latent} and FullyNAT \citep{gu2020fully}, are predominantly structured following the conventional framework described at the beginning of Section \ref{subsec:variationl_nmt}. Based on the latent variable design, FullyNAT utilizes the connectionist temporal classification (CTC) loss \citep{libovicky2018end} and achieves state-of-the-art (SOTA) performance in both translation quality and speed. 

On the other hand, there is non-latent variable based modeling. DA-transformer \citep{huang2022directed, ma2023fuzzy} has been proposed to predict directed acyclic graphical dependencies among target tokens, and PCFG-NAT \citep{gui2024non} generalizes the graphical dependencies to encompass non-adjacent and bidirectional structures. Although these recent works are also considered to achieve SOTA performance, they have not been directly compared to FullyNAT. Due to their reliance on tokenized BLEU scores \citep{papineni2002bleu}, which are dependent on specific preprocessing methods, direct comparison of reported BLEU scores is improper. Based on the reported improvements over their AT counterpart baselines, we carefully argue that FullyNAT attains superior translation quality and speed to the non-latent variable based methods. In this study, we apply a novel latent variable model to the existing superior latent variable modeling-based approaches, LaNMT and FullyNAT, elucidating the advantages of our proposed approach.

\section{Proposed Method}
\label{sec:proposed_method}
In this section, we describe the structural framework and objectives of our proposed novel latent variable model in Section \ref{subsec:dual_hierarchical_latent_variable_model}. Subsequently, we explain the design of each distribution, including likelihood, prior, and posterior, incorporating encoder and decoder parameters within the proposed LadderNMT model in Section \ref{subsec:application_to_nats}. Notably, our posterior estimation strategy is based on the ladder inference technique employed in LVAE. We elaborate on the advantages of this specific design in mitigating the OSPC problem. Lastly, we introduce a methodology for partially sharing the intermediate latent variable across both reconstruction processes in Section \ref{subsec:partial_latent_space_sharing}. We highlight the potential efficacy of partial sharing compared to entire sharing, particularly in scenarios where two languages necessitate a certain degree of distinct latent spaces.

\subsection{Dual Hierarchical Latent Variable Model}
\label{subsec:dual_hierarchical_latent_variable_model}
As previously outlined in the introduction, our proposed method is rooted in the concept of the dual reconstruction, which regards the translated sentence as a latent variable of the input sentence and conducts two simultaneous reconstruction processes. Additionally, we introduce an intermediate latent variable shared between the input and translated sentences, establishing a dual hierarchical latent variable model. This section describes the derivation of our approach's objective, beginning with the source reconstruction process. Since we focus on translation tasks, we extend this derivation to the supervised learning scenario, wherein the translated target sentence is provided by the dataset. In other words, the event of the translated target sentence of the source is sampled and fixed. Finally, we present the final objective, which comprises both source and target reconstruction objectives.

The objective of the source reconstruction process aims to maximize $\log{p(X)}$. Following the conventional derivation of variational inference \citep{bishop2006pattern} with two latent variables, $(Z,\hat{Y})$, where $\hat{Y}$ represents the translated target sentence, the objective is derived in the ELBO form:
\begin{align}
    \log{p(X)} 
    &\geq \int_{Z}\sum_{\hat{Y}}q(Z,\hat{Y}|X)\log{\left(p(X|Z,\hat{Y})\frac{p(Z,\hat{Y})}{q(Z,\hat{Y}|X)}\right)}dZ \nonumber \\
    &= \int_{Z}\sum_{\hat{Y}}q(Z|\hat{Y},X)q(\hat{Y}|X)\log{\left(p(X|Z,\hat{Y})\frac{p(Z|\hat{Y})p(\hat{Y})}{q(Z|\hat{Y},X)q(\hat{Y}|X)}\right)}dZ \nonumber \\
    &= \sum_{\hat{Y}}q(\hat{Y}|X)\int_{Z}q(Z|\hat{Y},X)\log{p(X|Z,\hat{Y})}dZ \nonumber \\
    &\quad + \sum_{\hat{Y}}q(\hat{Y}|X)\int_{Z}q(Z|\hat{Y},X)\log{\frac{p(Z|\hat{Y})}{q(Z|\hat{Y},X)}}dZ \nonumber \\
    &\quad + \sum_{\hat{Y}}q(\hat{Y}|X)\log{\frac{p(\hat{Y})}{q(\hat{Y}|X)}}\int_{Z}q(Z|\hat{Y},X)dZ. \nonumber
\end{align}
The first term of the objective formulation indicates the log-likelihood of observation marginalized by hierarchical latent variables. The second and third terms indicate KL divergences of the intermediate latent variable and the translated sentence, respectively. Given this derived objective, the supervised learning case can be obtained by replacing $\hat{Y}$ (the translated sentence) with $Y$ (real target sentence). Since $Y$ is the aligned latent variable of $X$, we eliminate the marginalization process for $\hat{Y}$. Based on Monte Carlo approximation and the assumption that there is no duplicated sentences in dataset, we assume $p(Y)$ and $p(X)$ are both uniform distributions with $1/N$ probability mass for every possible outcome, where $N$ is the number of data samples. Additionally, based on the assumption that the translation between true pair, $X$ and $Y$, is one-to-one mapping, we assume $q(Y|X)$ and $q(X|Y)$ are both 1. Finally, the objective is approximated in supervised learning as follows:
\begin{align}
    \mathcal{J}_{X}(X,Y) &\approx \int_{Z}q(Z|Y,X)\log{p(X|Z,Y)}dZ+\int_{Z}q(Z|Y,X)\log{\frac{p(Z|Y)}{q(Z|Y,X)}}dZ+\log{\frac{1}{N}} \nonumber \\
    &\approx \int_{Z}q(Z|Y,X)\log{p(X|Z,Y)}dZ+\int_{Z}q(Z|Y,X)\log{\frac{p(Z|Y)}{q(Z|Y,X)}}dZ \nonumber \\
    &=\mathbb{E}_{q(Z|Y,X)}[\log{p(X|Z,Y)}]-D_{KL}[q(Z|Y,X) \parallel p(Z|Y)]. \nonumber
\end{align}
The second approximation is done for simplicity because the constant, $\log{1/N}$, does not affect an optimal solution during optimization.
Analogously, the objective of target sentence reconstruction is derived as follow:
\begin{align}
    \mathcal{J}_{Y}(X,Y) &=\mathbb{E}_{q(Z|X,Y)}[\log{p(Y|Z,X)}]-D_{KL}[q(Z|X,Y) \parallel p(Z|X)]. \nonumber
\end{align}
Finally, the total objective of the dual hierarchical latent variable model is defined as:
\begin{align}
    \mathcal{J}(X,Y) &= \mathcal{J}_{X}(X,Y) +\mathcal{J}_{Y}(X,Y) \nonumber \\
    &=\mathbb{E}_{q(Z|X,Y)}[\log{p(X|Z,Y)}+\log{p(Y|Z,X)}] \nonumber \\
    &\quad -\alpha\left(D_{KL}[q(Z|X,Y) \parallel p(Z|X)]+D_{KL}[q(Z|X,Y) \parallel p(Z|Y)]\right). \label{eq:our_objective}
\end{align}
It is important to note that the intermediate latent variable $Z$ is sampled from the approximated posterior distribution $q(Z|X,Y)$ and is used for both likelihoods and KL divergences due to the latent space sharing.

\subsection{Application to NATs}
\label{subsec:application_to_nats}
We apply the dual hierarchical latent variable model to the NAT task. We are given two encoder-decoder models, $\theta=\{\theta^e,\theta^d,\theta^{pri}\}$ and $\phi=\{\phi^e,\phi^d,\phi^{pri}\}$, for source-to-target and target-to-source translations, respectively. $\theta^e$ and $\theta^d$ (or $\phi^e$ and $\phi^d$) refer encoder and decoder models' parameters, respectively. Usually, they are modeled by Transformer architecture \citep{vaswani2017attention}, except the causal mask design in decoder's self-attention layer to translate whole words at once. $\theta^{pri}$ (or $\phi^{pri}$) is weight parameters of linear layers that estimate the mean and variance of prior Gaussian distribution given the encoded representations computed by the encoder. Notably, there are no parameters for posterior estimations, which will be elaborated later. For the previous work, LaNMT \citep{shu2020latent}, there is an additional parameter set for length predictor that predicts the length of a translated sentence given an intermediate latent variable. On the other hand, FullyNAT \citep{gu2020fully} does not employ the length predictor, but simply predicts $U$ times bigger number of words compared to the input sentence length, then it simply eliminates tokens that are sequentially repeated. 
Our LadderNMT utilizes either the length prediction method from LaNMT or FullyNAT.

\subsubsection{Prior Distribution Design}
\label{subsubsec:prior_distribution_design}
The prior distribution of the source reconstruction process, denoted by $p(Z|X)$, is modeled by a Transformer encoder (including a word embedding layer \citep{mikolov2013efficient}) and linear layers parameterized by $\theta^e$ and $\theta^{pri}$. Initially, the Transformer encoder transforms the source sentence into deterministic hidden representations, expressed as $TE(X;\theta^e)=H^x=\{h^x_1,h^x_2,...,h^x_{T_x}\}$, where $TE$ represents the Transformer encoder function. Subsequently, each hidden representation undergoes linear transformations to estimate the mean and variance of Gaussian prior distributions, formulated as follows:
\begin{align}
    p(z_i|X;\theta^e,\theta^{pri})&=\mathcal{N}(z_i|\mu^x_i,\sigma^x_i), \quad 1\leq i \leq T_z \nonumber \\
    \mu^x_i&=LT(linear(H^x;\theta^{pri}_{\mu}),T_z)_i \in \mathbb{R}^{d_z}, \nonumber \\
    \sigma^x_i&=LT(softplus(linear(H^x;\theta^{pri}_{\sigma}),T_z)_i \in \mathbb{R}^{d_z}, \nonumber
\end{align}
where $softplus(a)=\log{(1+e^a)}$. 
Here, $LT(\cdot,L)$ denotes the length transformation function, which adjusts the length of the input vector sequence to a specified length $L$. As the intermediate latent space is shared for both languages' reconstructions, a unified sequence length $T_z$ is required for the latent variables. We adopt a monotonic location-based attention mechanism that incorporates input vectors with smoothly spread weights for the neighboring locations, as detailed in previous work \citep{shu2020latent}. For the LadderNMT application to the LaNMT model, a pre-defined value is set for $T_z$. Conversely, for application to the FullyNAT model, the length is dynamically transformed to $T_z=U(T_x+T_y)/2$.

Similarly, the prior distribution of the target reconstruction process, $p(Z|Y)$, is established following analogous procedures, with the input sentence $Y$ and parameters $\phi^e$ and $\phi^{pri}$ substituted accordingly.

\subsubsection{Posterior Distribution Design: Ladder Inference}
\label{subsubsec:posterior_distribution_design}
As mentioned briefly in Introduction, ladder inference involves the utilization of both inference and generation models to estimate the posterior distribution. In our proposed dual hierarchical latent variable model, the source-to-target $\theta$ and target-to-source $\phi$ models concurrently play the roles of inference and generation. Consequently, $\mu^x$ and $\sigma^x$, computed by $\theta$, are viewed as parameters estimated by the inference model, while $\mu^y$ and $\sigma^y$, computed by $\phi$, are viewed as parameters estimated by the generation model within the source reconstruction perspective. Subsequently, ladder inference is conducted to compute the posterior distribution according to the following formulation:
\begin{align}
    q(z_i|X,Y;\theta^e,\theta^{pri},\phi^e,\phi^{pri})&=\mathcal{N}(z_i|\mu^{xy}_i,\sigma^{xy}_i), \nonumber \\
    \mu^{xy}_i&=\frac{\mu^x_i \odot (\sigma^x_i)^{-1}+\mu^y_i \odot (\sigma^y_i)^{-1}}{(\sigma^x_i)^{-1}+(\sigma^y_i)^{-1}}, \nonumber \\
    \sigma^{xy}_i&=\frac{1}{(\sigma^x_i)^{-1}+(\sigma^y_i)^{-1}}, \nonumber
\end{align}
where $\odot$ indicates the element-wise multiplication operation.

The OSPC problem arises when the estimated posterior distribution tends to become independent of a conditional term. As illustrated in the example provided in Section \ref{subsec:variationl_nmt}, this occurs when the posterior estimator is sufficiently complex to disregard one side of the input information and there is an imbalanced importance between objective terms. Regarding the former factor, ladder inference exhibits a degree of resistance, as it incorporates the means and variances of both prior distributions through a simple dynamic. With the exception of extreme cases where a prior estimator assigns a zero mean or infinitely large variance, it is challenging to ignore one-sided information, contrasting with previous works' Transformer-based posterior models.

Furthermore, we assert that the structure of ladder inference prevents the OSPC problem stemming from the disproportionately large importance of KL divergences. To substantiate this claim, we derive the analytic solution of a KL divergence term, $D_{KL}[q(Z|X,Y) \parallel p(Z|Y)]$. Given that the posterior and prior distributions follow Gaussian families, the KL divergence term is derived as follows:
\begin{align}
    D_{KL}[q(Z|X,Y) \parallel p(Z|Y)]=\log{\frac{\sigma^y}{\sigma^{xy}}}+\frac{(\sigma^{xy})^2+(\mu^{xy}-\mu^{y})^2}{2(\sigma^y)^2}-\frac{1}{2}. \nonumber
\end{align}
The first and second derivatives of the KL divergence with respect to $\mu^x$ are derived as follows:
\begin{align}
    \frac{\partial D_{KL}}{\partial \mu^{x}} &=(\mu^{x}-\mu^{y})\frac{1}{(\sigma^{x}+\sigma^{y})^2}, \quad \frac{\partial^2 D_{KL}}{(\partial \mu^{x})^2}=\frac{1}{(\sigma^{x}+\sigma^{y})^2} \geq 0. \nonumber
\end{align}
Thus, it can be inferred that the optimal solution of the KL divergence is attained when $\mu^x$ equals $\mu^y$. Consequently, a high importance of the KL divergence term does not compel the model to disregard one side of the input information; rather, it regularizes the two distinct prior estimators to infer the same mean.

\begin{figure}
    \centering
    \includegraphics[width=0.9\linewidth]{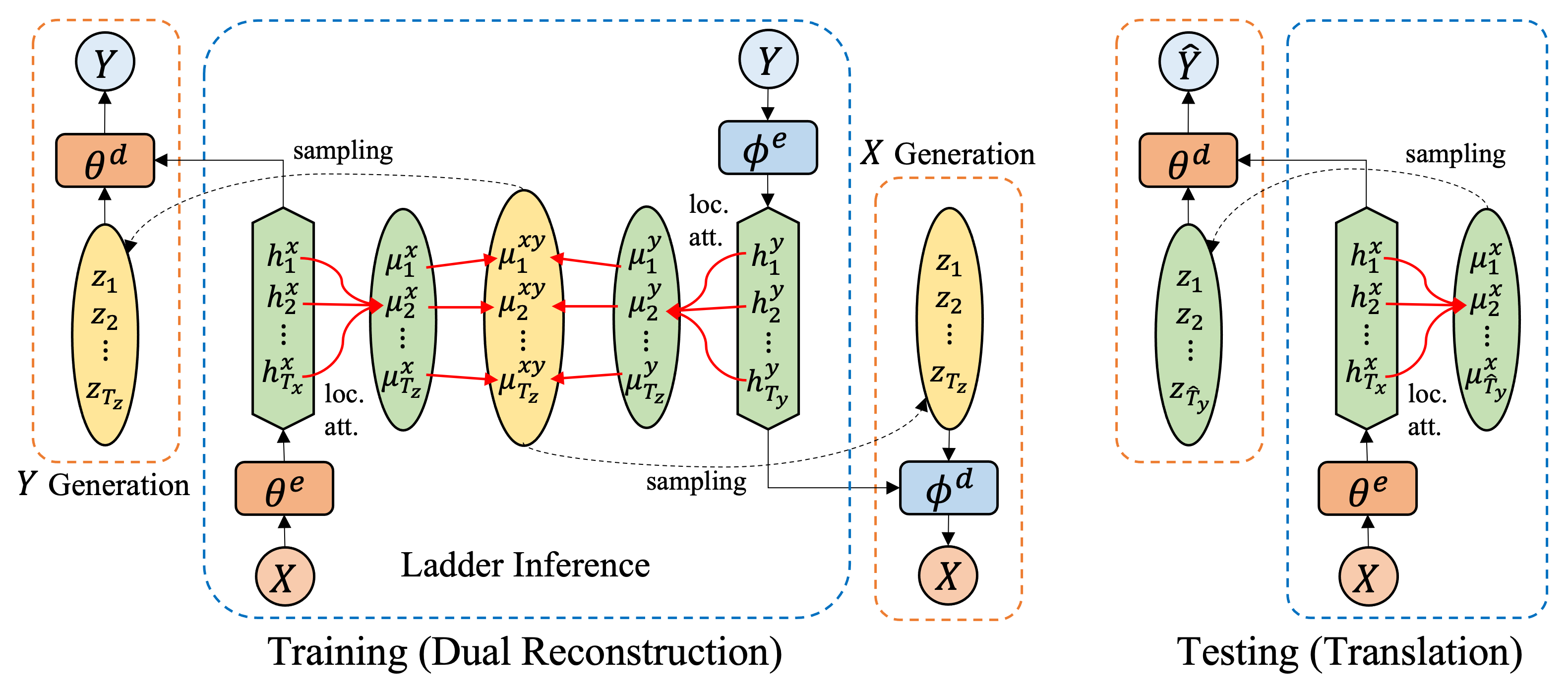}
    \caption{Illustrations of the training (left, dual reconstruction task) and testing (right, translation task) based on our proposed method, LadderNMT. We only illustrate the source-to-target translation task for the testing, the target-to-source translation task is done by the symmetric manner with the opposite model and input $Y$. $\mu$ terms represent each prior and posterior distributions. We omitted the element of variance parameters $\sigma$ from the prior and posterior distributions for simplicity. `loc. att.' means the monotonic location-based attention mechanism that transforms the total length.
    }
    \label{fig:ladder_nmt_architecture}
\end{figure}

\subsubsection{Likelihood Distribution Design}
\label{subsubsec:likelihood_distribution_design}
The likelihood distribution of the source reconstruction process, denoted as $p(X|Z,Y)$, is modeled using a Transformer decoder, parameterized by $\phi^d$, excluding the causal mask and the word embedding layer, and including the logit linear layer. This decoder takes as input the encoded hidden representations $H^y$ and latent variables $Z$ sampled from the posterior distribution $q(Z|X,Y)$ during training. During testing, $Z$ is sampled from the prior distribution $p(Z|Y)$. To enable backpropagation, the sampling process employs the reparameterization trick \citep{kingma2013auto}, defined as $z_i=\mu^{xy}_i+\pmb{\epsilon}\sigma^{xy}_i$ where $\pmb{\epsilon} \sim \mathcal{N}(\pmb{0},\pmb{I})$. The likelihood distribution is then computed as follows:
\begin{align}
    p(x_i|Z,Y;\phi^d)&=softmax(O^x_i), \nonumber \\
    O^x_i&=TD(Z,H^y;\theta^d)_i \in \mathbb{R}^{|V|}, \nonumber
\end{align}
where $TD$ represents the Transformer decoder function. Similarly, $p(Y|Z,X)$ for the target reconstruction process is computed using another Transformer decoder parameterized by $\theta^d$ with inputs $Z$ and $H^x$. It is important to note that the same $Z$ is used for both the source and target reconstruction processes due to the shared latent space.

The entire training (supervised learning) and testing processes of LadderNMT are illustrated in Fig. \ref{fig:ladder_nmt_architecture}. During training, at first, the two encoders compute the encoded hidden representations, $H^x$ and $H^y$. Second, the two prior estimators individually estimate the prior distributions. Third, the ladder inference process computes the posterior distribution from these prior distributions. Fourth, after sampling the intermediate latent variable, the two decoders generate $X$ and $Y$. Finally, the total objective of the dual hierarchical latent variable model, as given by Eq. \ref{eq:our_objective}, is computed, and the entire model is optimized jointly. During testing, we follow similar processes as in training, but the intermediate latent variable is drawn from the prior distribution. However, after the first iteration of translation, we can follow the ladder inference process and sample the intermediate latent variable from the posterior distribution given the first translation outcome in the setting of iterative refinement \citep{shu2020latent}.

\subsection{Partial Latent Space Sharing}
\label{subsec:partial_latent_space_sharing}
Instead of sharing the entire latent space in LadderNMT, we propose a variant that partially shares the intermediate latent space. This approach allocates subsets of dimensions to entirely focus on each language information. The proportion of shared versus non-shared dimensions is controlled by a predefined hyperparameter $\rho \in [0,1]$. The number of shared dimensions is calculated as $d^{srd}_z=round(d_z \times \rho)$, where $d_z$ is the dimension of the intermediate latent vector. Consequently, the number of non-shared dimensions is $d_z-d^{srd}_z$.

Partial sharing can be advantageous when the true manifolds of intermediate latent variables differ significantly between languages. Such differences may arise from substantial geographical, cultural, or historical disparities. For instance, the Inuktitut dialect from Nunavik (Arctic Québec) contains numerous expressions for `snow' \citep{2015inuktitut}, while languages from equatorial regions might have only a few terms for `snow'. This indicates that the latent spaces for these languages might diverge in specific aspects, such as vocabulary related to `snow'. In these scenarios, incorporating non-shared dimensions can help preserve these unique language characteristics.

\section{Non-Autoregressive Neural Machine Translation Results}

\subsection{Data Description}
\label{subsec:data_description}
Our experiments were conducted on three translation tasks: WMT14 English-German (En-De, 3.9M pairs), WMT18 English-Finnish (En-Fi, 2.9M pairs), and WMT18 English-Turkish (En-Tr, 0.2M pairs) \citep{bojar2014findings, rej2018findings}. We employed the same preprocessing, tokenization (including subword byte-pair encoding) methods used in the Fairseq toolkit \citep{ott2019fairseq}. For vocabulary sets, we selected the 32K, 10K, and 10K most frequent subwords from each training dataset, respectively. For validation, we used Newstest13 (3K pairs), Newstest16 (3K pairs), and Newsdev16 (1K pairs) for the En-De, En-Fi, and En-Tr tasks, respectively. For testing, we used Newstest14 (3K pairs), Newstest18 (3K pairs), and Newstest18 (3K pairs), respectively.

Sequence-level knowledge distillation (KD) \citep{kim2016sequence} has used for mitigating the multimodality problem in NAT training. It involves regenerating paired sentences using a pre-trained autoregressive translation (AT) model, thereby reducing one-to-many mapping examples. The bidirectional KD method further regenerates sentences using both source-to-target and target-to-source pre-trained AT models, producing two pairwise KD datasets from an original dataset \citep{ding2021rejuvenating}. In all our experiments, we employed the bidirectional KD method, as the dual reconstruction perspective inherently involves translations in both directions, thus necessitating two KD datasets. We utilized pre-trained AT models based on base Transformer architectures \citep{vaswani2017attention} to regenerate these KD datasets.

\subsection{Models and Training}
\label{subsec:models_and_training}
As discussed in Section \ref{subsec:non-autoregressive_nmt}, we applied our LadderNMT model to the architectures of LaNMT and FullyNAT (+VAE model \cite{gu2020fully}). We re-implemented LaNMT and FullyNAT based on our implementation of Transformer models, using their publicly available sources\footnote{https://github.com/zomux/lanmt}\footnote{https://github.com/shawnkx/Fully-NAT}. For the En-De and En-Fi experiments, we used the base Transformer architecture as detailed in \citep{vaswani2017attention}. For En-Tr, we employed a small Transformer architecture, which shares the base architecture's configurations except for a feed-forward layer dimension of 1024, 4 heads in multi-head attention layers, a learning rate of 0.0005, and a dropout rate of 0.3. The number of tokens per mini-batch was set to 8K, 4K, and 4K for En-De, En-Fi, and En-Tr, respectively. For all experiments, we used an inverse square root learning rate scheduler and checkpoint ensemble with the latest 10 checkpoints \citep{ott2019fairseq}. The best checkpoint was saved based on validation results, and early stopping was implemented when validation results did not surpass previous best results for 50 patience steps.

Apart from the above basic configurations, we adhered to the same configurations for LaNMT and FullyNAT related to NAT approaches \citep{shu2020latent, gu2020fully}, such as the auxiliary posterior model configurations, length prediction method, intermediate latent space dimension, annealing schedule of $\alpha$, and upsample ratio of the latent variable (for FullyNAT). Notably, we set the default $\alpha$ to 0.5 for the FullyNAT baseline, which was optimal based on our internal ablation studies. A significant deviation from the FullyNAT baseline configuration was the number of tokens per mini-batch: we used 8K for En-De while they used 128K. This adjustment was necessary due to computation and memory resource constraints. Note that FullyNAT's upsampling operation substantially increases computation and memory costs three times for the decoder part, and the CTC loss also increases such costs for its marginalization with respect to possible paths derived by the dynamic programming algorithm. Consequently, our re-implementation of the FullyNAT baseline showed approximately a 1.0 BLEU score point degradation compared to the original paper. However, our application of LadderNMT to the FullyNAT architecture used consistent configurations for fair comparison.

Regarding the configurations of our LadderNMT applications, we generally followed the baseline configurations. However, we empirically identified optimal configurations for the intermediate latent space dimension and KL divergence coefficient ($d_z$/$\alpha$): (16/1.5) for En-De, (32/1.5) for En-Fi, and (32/1.0) for En-Tr. We also varied the partial sharing ratio $\rho$ and selected the best values: 0.25 for En-De, 0.25 for En-Fi, and 1.00 for En-Tr. Notably, our LadderNMT models achieved their best performance with $d_z$ values larger than the baselines' optimal configuration of $d_z=8$, suggesting that LadderNMTs' shared intermediate latent spaces can capture more independent information compared to the baselines. Additionally, we found that $\alpha$ values exceeding the baselines' optimal $\alpha=1.0$ (and 0.5 for FullyNAT) corroborate our argument regarding the regularization benefit of KL divergence, as discussed in Section \ref{subsubsec:posterior_distribution_design}.

All En-De experiments were conducted on two NVIDIA RTX4090 GPUs, averaging around 240 hours of training. For En-Fi and En-Tr tasks, we used two NVIDIA GTX1080Ti GPUs, with an average training time of approximately 220 hours.

\subsection{Decoding and Evaluations}
\label{subsec:evaluations}
The decoding process for models based on the LaNMT architecture involved three iterative refinement steps. For models based on the FullyNAT architecture, sequentially repeated tokens were simply eliminated. In the case of autoregressive Transformer baselines, we utilized the beam search algorithm with a beam width of 5.

To evaluate translation quality, we employed the case-insensitive detokenized BLEU metric \citep{papineni2002bleu} for models based on the LaNMT architecture, adhering to the original paper's evaluation metric\footnote{Although their paper describes SacreBLEU \citep{post2018call} as the evaluation metric, their source code seems to use the case-insensitive detokenized BLEU with the `intl' tokenizer.}. For models based on the FullyNAT architecture, we followed the original paper's metric, tokenized BLEU. Additionally, we reported the average translation speed (in milliseconds) measured on each testset using a single GTX1080Ti GPU. Lastly, we compared the total number of parameters across the models.

\begin{table}[h]
\caption{Experiment results for models based on the LaNMT architecture. $P$ means the total number of parameters in the unit of million(M). The left and right numbers of the `/' at the columns of each testset represent translation quality (BLEU scores) of English-to-\textit{language} and of \textit{language}-to-English, respectively. Translation speed(Time) is measured in milliseconds (ms). The term '+Layers' indicates LadderNMT models with an increased number of layers to match the similar number of parameters compared to LaNMT baselines: specifically, we added 5, 3, and 4 layers for the En-De, En-Fi, and En-Tr tasks, respectively.
}
\label{table:lanmt_results}
\begin{tabular*}{1.0\textwidth}{@{\extracolsep{0.0in}}l|ccc|ccc|ccc}
\toprule
& \multicolumn{3}{|c|}{WMT14 En-De}
& \multicolumn{3}{|c|}{WMT18 En-Fi}
& \multicolumn{3}{|c}{WMT18 En-Tr}
\\ \cmidrule{2-4}\cmidrule{5-7}\cmidrule{8-10}%
Model & $P$ & News14 & Time & $P$ & News18 & Time & $P$ & News18 & Time \\
\midrule
Transformer  & 154 & 26.5/32.0 & 162 & 109 & 14.1/20.0 & 191 & 64 & 16.0/19.5 & 196\\
LaNMT   & 232 & 23.8/30.2 & 22 & 164 & 11.6/17.7 & 13 & 98 & 12.5/16.3 & 14\\
w/ LadderNMT   & 155 & 24.9/30.4 & 28 & 109 & 12.3/17.7 & 17 & 66 & 12.6/16.4 & 15\\
+ Layers   & 229 & 25.5/31.2 & 38 & 154 & 12.4/17.9 & 24 & 95 & 12.9/16.9 & 23\\
\botrule
\end{tabular*}
\end{table}

\begin{table}[h]
\caption{Experiment results for models based on the FullyNAT architecture. The left and right numbers of the `/' at the columns of each testset represent translation quality (BLEU scores) of English-to-\textit{German} and of \textit{German}-to-English, respectively.}
\label{table:fullynat_results}
\begin{tabular}{@{}l|c|c|c@{}}
\toprule
& \multicolumn{3}{@{}c@{}}{WMT14 En-De}
\\\cmidrule{2-4}%
Model & Param(M) & News14 & Time(ms)  \\
\midrule
Transformer  & 153 & 27.6/33.6 & 162 \\
FullyNAT   & 166 & 26.2/30.4 & 6 \\
w/ LadderNMT   & 121 & 26.9/31.5 & 7 \\
\botrule
\end{tabular}
\end{table}

\subsection{Translation Results}
\label{subsec:translation_results}
Table \ref{table:lanmt_results} presents the experiment results of models based on the LaNMT architecture. Compared to LaNMT, our application of LadderNMT to the LaNMT baseline (`w/ LadderNMT') significantly reduced the number of parameters while achieving comparable or superior (En-De task) BLEU scores. Although there is a slight reduction in translation speed compared to LaNMT, it remains significantly faster than the autoregressive Transformer (AT) baseline. Additionally, when we increased the number of layers in LadderNMT to match the number of parameters in LaNMT, LadderNMT\textit{+Layers} significantly outperformed LadderNMT for the WMT14 En-De task that has a relatively large dataset. We believe that our proposed approach has an efficient architecture with a less number of parameters, so that it can scale-up other parts of the model to improve the performance. 

Table \ref{table:fullynat_results} displays the WMT14 En-De experiment results for models based on the FullyNAT architecture, a state-of-the-art NAT model. Notably, our adaptation of LadderNMT to the FullyNAT architecture (`w/ LadderNMT') resulted in improved BLEU scores for both the English-to-German and German-to-English tasks, while reducing the total number of parameters by approximately 27\%. Furthermore, this adaptation does not significantly degrade translation speed compared to the FullyNAT baseline. These results suggest that our approach can reduce the number of parameters in the existing NAT models, while preserving (or improving) translation quality and speed. 

\section{Qualitative Analyses of the Shared Intermediate Latent Space}
\label{sec:anaylsis_of_latent_space}
 To understand the impact of our proposed approach on the shared intermediate latent space, we conducted qualitative analyses of this space. Hereafter, we refer to the intermediate latent variable simply as the latent variable for brevity. For the analyses detailed in Sections \ref{subsec:2dim_visualizations} and \ref{subsec:cca_analysis}, we utilized pre-trained LaNMT and LadderNMT (applied to LaNMT) models on the En-De task. Latent variables were collected using Newstest14. Specifically, the latent variable $Z$ was drawn from the prior distributions, $p(Z|X)$ and $p(Z|Y)$. In other words, the latent variables collected for LadderNMT were independently computed by the two encoder and prior estimators with their parameters ($\theta^{e},\theta^{pri}$) and ($\phi^{e},\phi^{pri}$), reflecting the testing scenario. For the OSPC analysis detailed in Section \ref{subsec:ospc_study}, we collected $Z$ sampled from the posterior distribution, $q(Z|X,Y)$.

\begin{figure}[]
    \centering
    \hbox{\centering \hspace{0.8in} \includegraphics[width=0.6\linewidth]{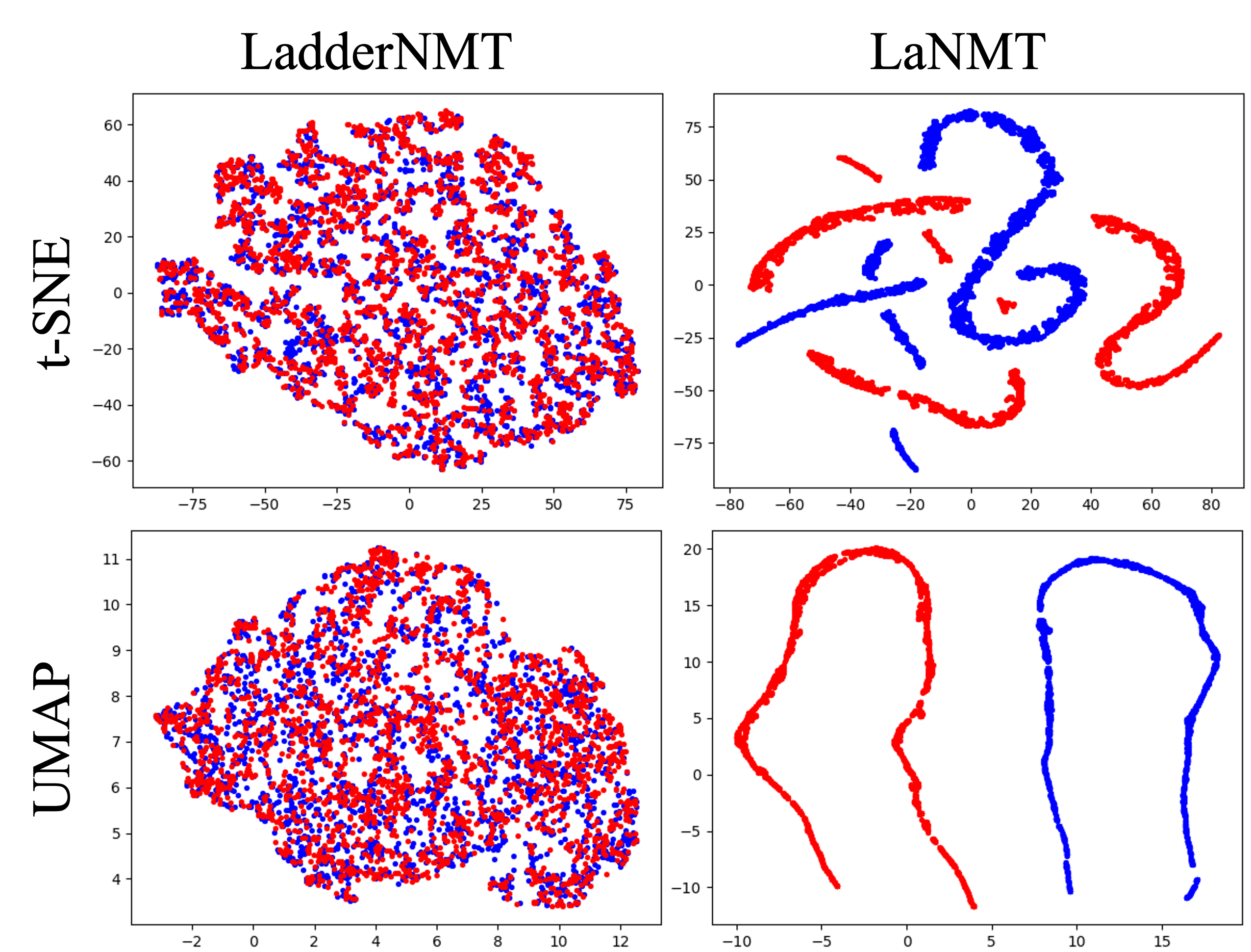} }
    \caption{Latent variables of LadderNMT and LaNMT in 2-dimensional space by t-SNE and UMAP. Red and blue dots are latent variables from English and German sentences, respectively.}
    \label{fig:2d_visualization}
\end{figure}

\subsection{2-Dimensional Visualizations}
\label{subsec:2dim_visualizations}
First, we visualized the collected latent variables of LadderNMT and LaNMT in two-dimensional space using t-SNE \citep{van2008visualizing} and UMAP \citep{mcinnes2018umap}, as depicted in Fig. \ref{fig:2d_visualization}. Red dots represent latent variables for English, while blue dots represent those for German. The latent variables of LaNMT are clearly distinguishable by language. It is important to note that we suggest that LaNMT figures as a comparable baseline representing two individually trained latent spaces, not that these latent variables contain disadvantageous language-specific information for translation tasks. Conversely, LadderNMT's latent variables are indistinguishable by language, which suggests that latent space sharing has a regularization effect, promoting the extraction of language-agnostic information rather than language-specific information.

\subsection{Canonical Correlation Analysis (CCA)}
\label{subsec:cca_analysis}
While Section \ref{subsec:2dim_visualizations} demonstrated that LadderNMT extracts language-agnostic information, it remains to be verified whether the latent variables from paired sentences are closely mapped, as posited in Section \ref{subsubsec:posterior_distribution_design} (optimal solution derivation of KL divergence in LadderNMT). In this section, we examine the similarity of paired sentences in the latent space. We employ canonical correlation analysis (CCA) \citep{hotelling1936cca} and measure the CCA prediction score to assess the similarity of paired sentences in the latent space. The CCA method involves two trainable linear transformation layers for the two paired input modalities, which are trained to maximize the canonical correlation between paired data in a lower-dimensional space (output space of the linear layers). The prediction score is formulated as follows: $\left(1-\frac{(\hat{Z}_X-\hat{Z}_Y)^2}{(\hat{Z}_X-m(\hat{Z}_X))^2}\right)$, where $m(\hat{Z}_X)$ is the mean of all $\hat{Z}_X$. Here, $\hat{Z}_X$ and $\hat{Z}_Y$ are the paired English and German latent variables whose dimensions are reduced by the linear transformations. A high CCA prediction score indicates that the latent variables of paired sentences are mapped closely to each other.

The CCA parameters were trained using the collected latent variables from the validation set (Newstest13) and subsequently applied to the collected latent variables from Newstest14. LadderNMT achieved a score of \textbf{0.6851}, compared to \textbf{0.2836} for LaNMT. Together with the analysis from Section \ref{subsec:2dim_visualizations}, this indicates that LadderNMT not only extracts language-agnostic information but also maps paired sentences closely together. We attribute this to the regularization effect of the KL divergences in our objective function, as defined in Eq. \ref{eq:our_objective}, which effectively align the two encoders and their prior estimations.

\begin{figure}[]
    \hbox{\centering \hspace{0.8in}
    \includegraphics[width=0.7\linewidth]{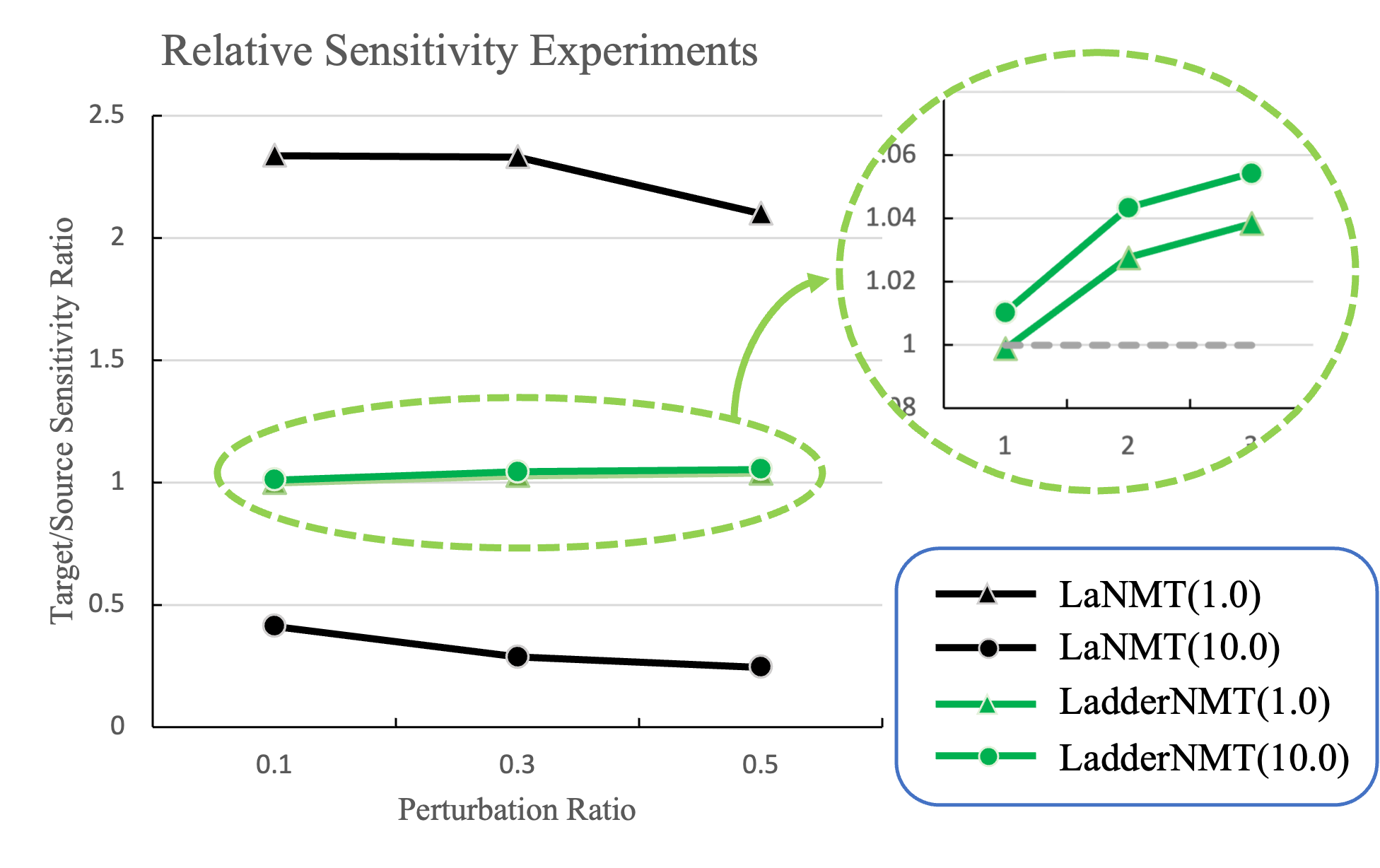} }
    \caption{Relative sensitivity test result. Each number in the parenthesis next to the model label is the KL coefficient.}
    \label{fig:relative_sensitivity_graph}
\end{figure}

\subsection{Relative Sensitivity for OSPC Study}
\label{subsec:ospc_study}
As a measure of the OSPC problem, we propose a new method, `relative sensitivity' test, to assess the model's relative sensitivity to input noise added to source or target sentences. In this test, we randomly altered several words in the source or target sentence alternatively (not simultaneously) and calculated the Euclidean distance between the latent variables inferred from the perturbed and original sentences. A model without OSPC issues would exhibit similar sensitivities to perturbations in both source and target inputs, resulting in a relative sensitivity ratio close to 1.

To clearly verify the existence of the OSPC problem in the original ELBO and our objective function, Eqs. \ref{eq:elbo_original} and \ref{eq:our_objective}, we trained LaNMT baseline models and our LadderNMT models using two different KL divergence coefficients $\alpha$: 1.0 and 10.0. Figure \ref{fig:relative_sensitivity_graph} illustrates the relative sensitivities of LaNMT and LadderNMT with these different KL coefficients. As anticipated, LaNMT's relative sensitivity significantly deviates from the 1.0 line, indicating that its posterior model neglects one side of the information. In contrast, LadderNMT's relative sensitivity remains close to 1.0, regardless of the KL coefficients. This result corroborates our assertion in Section \ref{subsubsec:posterior_distribution_design} that LadderNMT is not prone to the OSPC problem related to the importance of the KL divergence terms.


\section{Conclusion}
We proposed a novel latent variable modeling approach for non-autoregressive neural machine translation, grounded in a different perspective on the translation task and an advanced hierarchical latent variable model. Unlike conventional latent variable modeling, our approach regularizes the latent space to extract informative features from the input sentence without increasing the parameters of the posterior estimation model. Additionally, it motivates that the posterior estimation remains equally sensitive to all parts of the input information.

Our quantitative experiment results demonstrate that the proposed approach achieves superior translation quality compared to several baselines, including a state-of-the-art model. Moreover, the qualitative analyses of the latent space confirm that LadderNMT infers more informative latent variables that are language-agnostic and resistant to the issue of posterior estimation insensitivity to parts of the input information.

For future research, we intend to apply our proposed method to multilingual neural machine translation since we believe that the regularization effects by sharing the latent space can be amplified when multiple languages are involved. Also, our approaches are applicable to other tasks that have structural duality, such as text-to-image, image-to-text, text-to-speech and speech-to-text generation tasks. In addition, our approaches can be applied to non-latent variable-based models, such as DA-Transformer, with equipping the latent variable architecture.

\backmatter





\bmhead{Acknowledgements}
This research was supported by Basic Science Research Program through the National Research Foundation of Korea funded by the Ministry of Education (NRF-2022R1A2C1012633)

\section*{Declarations}
DongNyoneg Heo and Heeyoul Choi report financial support was provided by National Research Foundation of Korea.


\bibliography{sn-bibliography}

\end{document}